%% file: main.tex
\documentclass[runningheads]{llncs}
\usepackage[utf8]{inputenc} % allow utf-8 input
\usepackage[T1]{fontenc}    % use 8-bit T1 fonts
\usepackage{hyperref}       % hyperlinks
\usepackage{url}            % simple URL typesetting

\usepackage{amsfonts}       % blackboard math symbols
\usepackage{nicefrac}       % compact symbols for 1/2, etc.
\usepackage{microtype}      % microtypography
\usepackage[dvipsnames]{xcolor}
\usepackage{orcidlink}
\usepackage{amssymb}
\usepackage{algorithm}
\usepackage{algpseudocode}
\usepackage{subcaption}
\usepackage{amsmath}
\usepackage{rotating}
\usepackage{makecell}
\usepackage{multirow}
\usepackage{enumitem}
\usepackage{tikz}
\usetikzlibrary{tikzmark}
\usepackage{booktabs} 
\usepackage{graphicx,verbatim}
\usepackage{arydshln}
\usepackage{tabularx} % also loads 'array' package
\usepackage{tabulary}
\usepackage{colortbl}
\begin{document}
\title{MoSFormer: Augmenting Temporal Context with Memory of Surgery for Surgical Phase Recognition}
\titlerunning{MoSFormer}
%
\begin{comment}  %% Removed for anonymized MICCAI 2025 submission
\author{Hao Ding\inst{1} 
\and Xu Lian\inst{1} 
\and Mathias Unberath\inst{1}}
%
\authorrunning{H. Ding et al.}
\institute{Johns Hopkins University, Baltimore MD 21218, USA 
\email{hding15@jhu.edu,unberath@jhu.edu} \\}
\end{comment}

\author{Hao Ding\inst{1} 
\and Xu Lian\inst{1} 
\and Mathias Unberath\inst{1}}
\authorrunning{H. Ding et al.}
\institute{Johns Hopkins University, Baltimore MD 21218, USA 
\email{hding15@jhu.edu,unberath@jhu.edu} \\}

% \author{Anonymized Authors}  %% Added for anonymized MICCAI 2025 submission
% \authorrunning{Anonymized Author et al.}
% \institute{Anonymized Affiliations \\
%     \email{email@anonymized.com}}

\maketitle              % typeset the header of the contribution
\begin{abstract}
Surgical phase recognition from video enables various downstream applications. Transformer-based sliding window approaches have set the state-of-the-art by capturing rich spatial-temporal features. However, while transformers can theoretically handle arbitrary-length sequences, in practice they are limited by memory and compute constraints, resulting in fixed context windows that struggle with maintaining temporal consistency across lengthy surgical procedures. This often leads to fragmented predictions and limited procedure-level understanding.
To address these challenges, we propose Memory of Surgery (MoS), a framework that enriches temporal modeling by incorporating both semantic interpretable long-term surgical history and short-term impressions. MoSFormer, our enhanced transformer architecture, integrates MoS using a carefully designed encoding and fusion mechanism. We further introduce step filtering to refine history representation and develop a memory caching pipeline to improve training and inference stability, mitigating shortcut learning and overfitting.
MoSFormer demonstrates state-of-the-art performance on multiple benchmarks. On the Challenging BernBypass70 benchmark, it attains 88.0 video-level accuracy and phase-level metrics of 70.7 precision, 68.7 recall, and 66.3 F1 score, outperforming its baseline with 2.1 video-level accuracy and phase-level metrics of 4.6 precision, 3.6 recall, and 3.8 F1 score. Further studies confirms the individual and combined benefits of long-term and short-term memory components through ablation and counterfactual inference. Qualitative results shows improved temporal consistency. The augmented temporal context enables procedure-level understanding, paving the way for more comprehensive surgical video analysis.

%Furthermore, an ablation study confirms the individual and combined benefits of long-term and short-term memory components. Qualitative results show improved temporal consistency and procedure-level understanding. The counterfactual inference experiment further validates the contribution by the augmented temporal context.
% MoS extend the temporal context of transformer-based architectures, paving the way for more comprehensive surgical video analysis.

\keywords{Surgical workflow analysis \and Surgical data science  \and Surgical video analysis \and Surgical video processing}
% Authors must provide keywords and are not allowed to remove this Keyword section.

\end{abstract}

\input{sections/1_intro}    
\input{sections/2_method}

\input{sections/3_experiment}
\input{sections/4_conclusion}

\begin{credits}
\subsubsection{\ackname}
This research is supported by a collaborative research agreement with the MultiScale Medical Robotics Center at The Chinese University of Hong Kong. 

% \subsubsection{\discintname}
% It is now necessary to declare any competing interests or to specifically
% state that the authors have no competing interests. Please place the
% statement with a bold run-in heading in small font size beneath the
% (optional) acknowledgments\footnote{If EquinOCS, our proceedings submission
% system, is used, then the disclaimer can be provided directly in the system.},
% for example: The authors have no competing interests to declare that are
% relevant to the content of this article. Or: Author A has received research
% grants from Company W. Author B has received a speaker honorarium from
% Company X and owns stock in Company Y. Author C is a member of committee Z.
\end{credits}

\bibliographystyle{splncs04}
\bibliography{main}
\end{document}

%% file: sections/1_intro.tex
\section{Introduction}
\label{section:intro}

% \begin{figure}[t]
%   \centering
%    \includegraphics[width=1.0\textwidth]{figures/intro.pdf}
%    \caption{Illustration of the memory-based augmentation for the current sliding window-based surgical phase recognition paradigm. Existing approaches rely on a sliding window for phase prediction, disregarding the rich temporal context in surgical videos. Our MoS-based framework captures temporal information through long-term history and short-term impressions, integrating them into existing architectures to augment temporal context in surgical video analysis.}
%    \label{fig:intro}
% \end{figure}

Surgical phase recognition (SPR) is a vital aspect of surgical video analysis, providing essential procedure-level insights for applications such as video summarization, skill assessment, and intervention assistance. Surgical videos typically capture continuous, lengthy procedures with strong temporal dependencies, making temporal modeling a crucial challenge.
Traditional SPR approaches modeled surgical workflows as finite-state machines, predominantly using Hidden Markov Models (HMMs)\cite{BlumPFN08HMM,BhatiaOXH07RL_video_hmm,CadeneRTC16CNN_HMM_SMOOTHING,PadoyBFBN08_online,BlumFN10,PadoyBAFBN12}. However, these methods relied on handcrafted features and struggled with capturing complex visual and temporal dependencies. 
The advent of deep learning, particularly convolutional neural networks (CNNs), significantly improved feature extraction and SPR performance when leveraged with temporal models such as hierarchical HMMs, long short-term memory (LSTM) networks, and temporal convolutional networks (TCNs)~\cite{TwinandaSMMMP17endonet,JinLDCQFH20mrcnet,JinD0YQFH18sv-rcnet,YiJ19hfom,GaoJDH20treesearch,czempiel2020tecno,LeaVRH16tcn,FarhaG19mctcn,HochreiterS97lstm}. These temporal modeling methods suffer from either vanishing gradients or limited receptive fields when the sequence becomes lengthy. 

More recently, Transformers~\cite{VaswaniSPUJGKP17transfomer} and Vision Transformers (ViTs)\cite{DosovitskiyB0WZ21vit} have redefined sequence modeling by enabling parallelized computation and capturing long-range dependencies. Several transformer-based architectures have been introduced for SPR\cite{valderrama2020tapir,GaoJLDH21transsvnet,yang2024surgformer,KilleenZMATOU23_Pelphix,ding2024towards}, primarily in a sliding-window manner due to transformer's overwhelming memory consumption. This paradigm has achieved strong performance due to the superior extraction of spatial-temporal features of ViTs. However, the sliding-window approach inherently limits temporal modeling, as its fixed window size prevents the model from capturing long-horizon dependencies and procedure-level context. 
Various solutions have been proposed to address this problem, including hierarchical temporal modeling with two-stage training\cite{liu2023skit,Liu23Lovit}, temporal downsampling\cite{yang2024surgformer}, and finite-state machine modeling~\cite{ding2024neural}. 
%We argue that a way of directly capture and integrate procedure-level temporal information will also be helpful.

Memory mechanisms have long been applied for sequence processing from long short term memory~\cite{HochreiterS97lstm} to neural turing machine~\cite{graves2014neural}. Its recent success in medical image analysis~\cite{shen2023movit} and video object segmentation~\cite{Oh_2019_ICCV,cheng2021rethinkingspacetimenetworksimproved,cheng2022xmemlongtermvideoobject,cheng2024puttingobjectvideoobject}, indicates that it can be a promising avenue to accomplish our goal. Indeed, prior work, TMRNet~\cite{JinLCZDH21tmrnet} which used an LSTM temporal model, applied the memory mechanism in latent space to alleviate the vanishing gradients in LSTM, achieving great success. However, directly applying such memory-based approaches to current ViT architectures presents unique challenges. Unlike segmentation tasks, which rely on detailed spatial correspondences, SPR involves low-dimensional categorical labels that can lead to shortcut learning and overfitting when memory is applied naively in a sequential training fashion.

\begin{figure}[t]
  \centering
   \includegraphics[width=1.0\textwidth]{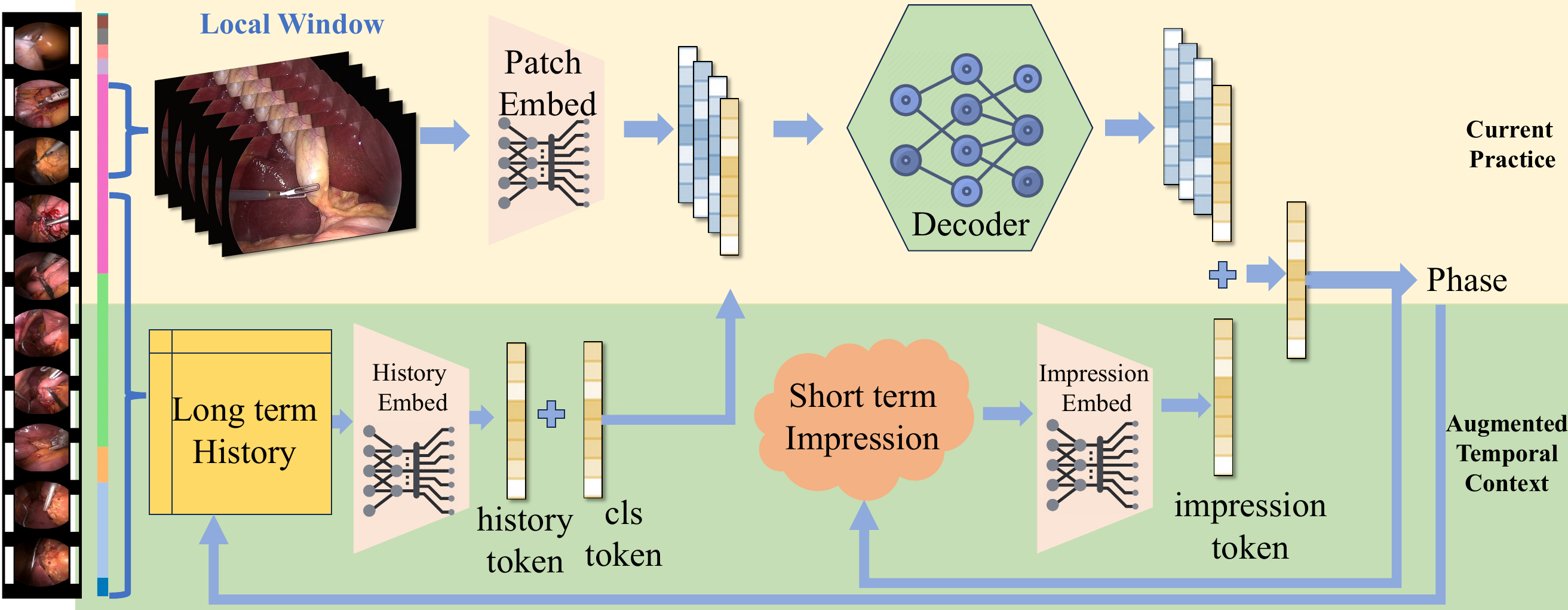}
   \caption{Illustration of the memory-based augmentation for the current sliding window-based surgical phase recognition paradigm. Existing approaches rely on a sliding window for phase prediction, disregarding the rich temporal context in surgical videos. Our MoS-based framework captures temporal information through long-term history and short-term impressions, integrating them into existing architectures to augment temporal context in surgical video analysis.}
   \label{fig:architecture}
\end{figure}

In this regard, we introduce Memory of Surgery (MoS), a complementary memory mechanism to augment the temporal context for surgical video analysis for the state-of-the-art sliding-window paradigm. MoS consists of a long-term history, which captures semantically understandable surgical history to represent the entire procedure, and a short-term impression, which captures the visual features of the previous video window to expand the receptive field. To effectively integrate MoS into existing architectures, we develop memory encoding and fusion mechanisms that encode historical information into the feature space and seamlessly fuse it with the spatial-temporal representation of the current clip. Additionally, to mitigate potential overfitting and shortcut learning, we introduce step filtering for history representation and propose a memory caching pipeline, enabling end-to-end training without the capacity loss observed in two-stage hierarchical architectures. All these innovations are realized in MoSFormer, one promising example of a transformer-based surgical phase recognition model that integrates MoS and the corresponding encoding, fusion, and memory caching mechanisms into a contemporary backbone, namely Surgformer~\cite{yang2024surgformer}. Our experiments on public benchmarks-Cholec80~\cite{TwinandaSMMMP17endonet}, AutoLaparo~\cite{wang2022autolaparo}, and BernBypass70~\cite{Lavanchy2024}—demonstrate the effectiveness of the proposed method, resulting in state-of-the-art performance. Especially on the challenging public benchmarks BernBypass70~\cite{Lavanchy2024}, MoSFormer achieves remarkable video-level accuracy of 88.0 and phase-level F1 scores of 66.3, improving over its baseline Surgformer~\cite{yang2024surgformer} by 2.1 and 3.8. Our ablation study and counterfactual inference experiment further validates the effectiveness of both short-term and long-term memory components and qualitative visualizations illustrate more temporally consistent predictions. 
%These results validate that MoS effectively extends the temporal receptive field of transformer-based architectures, enhancing the temporal context of current sliding window methods.

The primary contributions of our work are summarized as:
\begin{itemize}[topsep=0pt]
\item We introduce memory of surgery (MoS) along with its memory encoding, fusion, and caching mechanisms, to provide long-horizon information and overcome the temporal constraints of state-of-the-art architectures.
\item We propose MoSFormer, a realization of MoS, which achieves state-of-the-art results on multiple surgical phase recognition benchmarks.
\end{itemize}

%% file: sections/2_Method.tex
\section{Method}
\label{section:method}
We present our approach in two parts. First, we define Memory of Surgery (MoS), detailing its representation for long-term history and short-term impression. Second, we describe the MoSFormer architecture, including the encoding and fusion mechanisms that incorporate MoS and the memory-caching training and inference pipeline, which ensures effective and stable memory utilization during model learning and deployment.

\subsection{\bf{Memory of Surgey (MoS)}}
%Xu write here the representation format of long term history and short term feature
\noindent\textbf{Long-Term History.}
 To capture phase-level semantics over the entire procedure, we represent long-term history as entries of accumulated phase history. Each phase history entry consists of a one-hot encoded vector that captures phase identity, and a discretized duration measure through step filtering, where the step number is calculated by $step\_number = \lfloor \frac{num\_frames}{step\_size} \rfloor$. This step filtering has two purposes: 
 % (1) \textbf{Outlier Suppression During Inference:} By discarding small variations at the frame level, the model is protected from noise arising from very brief or anomalous phase segments during inference. 
 % (2) \textbf{Avoiding Over-Specificity:} By coarsening precise durations, the model focuses on general temporal patterns rather than memorizing specific counts, mitigating overfitting risks.
\begin{itemize}[leftmargin=*, label=\textbf{$\bullet$}]
    \item \textbf{Outlier Suppression During Inference:} By discarding small variations at the frame level, the model is protected from noise arising from very brief or anomalous phase segments during inference.
    \item \textbf{Avoiding Over-Specificity:} By coarsening precise durations, the model focuses on general temporal patterns rather than memorizing specific counts, mitigating overfitting risks.
\end{itemize}
% Importantly, the history tokens are arranged in the dataset’s default phase order rather than the exact chronological sequence of the target video, further reducing the chance of overfitting.
Each history entry also includes a binary masking variable to indicate whether a phase has presented in the current video. A complete history entry is thus structured as: $[\,0,\ldots,1,0,\ldots,\text{num\_steps},\text{mask}(\{0,1\})\,]$

\noindent\textbf{Short-Term Impression.}
While long-term history provides procedure-level context, the short-term impression module caches recent visual context to augment local temporal coherence. Specifically, we store final classifier (cls) tokens from previously processed frames and retrieve them when processing following frames. 
%At each time step: (1) The model processes a frame and extracts its cls token from the final transformer layer. (2) This cls token is stored in a cache, maintaining a rolling memory of recent frames. (3) When processing a new frame, the cached cls tokens from past frames are retrieved and integrated alongside the long-term history. This design allows the model to extend its receptive field beyond the fixed input window, leveraging both global procedural trends (via long-term history) and localized visual cues (via short-term impressions) for robust and context-aware phase recognition.

\subsection{\bf{MoSFormer}}
\noindent\textbf{Architecture.} The MoSFormer architecture, illustrated in Figure~\ref{fig:architecture}, integrates MoS within a ViT-based phase recognition framework. A sliding window with a specific length (16 in this practice) is captured and processed via patch embedding to form spatial temporal tokens. The long-term history is processed via a history embedding into a history token, which is then element-wise added to the trainable cls token. The history-augmented cls token is concatenated with the spatial temporal tokens and fed into the decoder blocks of a ViT-based architecture. The short-term impression is processed via an impression embedding into an impression token and element-wise added to the output cls token from the decoder. The final augmented cls token is used to predict the phase of the last frame in the sliding window. The final augmented cls token and predicted phase results are stored in short-term impression and long-term history.

\noindent\textbf{Encoding and Fusion.}
We encode both the long-term history tokens and short-term impression tokens via two-layer multi-head self-attention blocks. For the history tokens, the one-hot encoded phases are projected into embedding space, and a sinusoidal positional embedding is added to encode each phase step number. The embedded history entries are then appended by a history token and fed into self-attention blocks. The processed history token from the transformer is then early fused with the cls token via element-wise addition. The cached short-term impression tokens are appended with a learnable impression token and similarly undergo self-attention blocks. This processed impression token is subsequently late fused with the cls token via element-wise addition, yielding a final embedding that incorporates both long-range context and immediate past observations. This dual-stream fusion mechanism allows MoSFormer to jointly capture surgical progression trends while maintaining local phase consistency.%, enhancing both stability and adaptability.

\noindent\textbf{Memory-caching Training and Inference.}
For the history component, during training, history tokens are directly extracted from ground-truth phase annotations. During inference, history tokens are dynamically updated using the model’s predicted phases, ensuring an adaptive and evolving history representation.
For the short-term impression, previous works applied sequential training pipelines that continuously accumulate and consolidate memory to capture all preceding information. Directly applying this pipeline, however, will potentially result in severe overfitting as the the model will be sequentially trained on batches of the same labels given the long duration of the phases. Instead, we retain the randomized training pipeline but maintain a local cache for each frame during training and update each frame’s impression token once that the frame is processed in each batch. This approach allows the model to use the impressions from the previous epoch, ensuring the impression cache gradually converges alongside the model itself. During inference, we simply store and retrieve impressions from frames already processed. Since no random batch ordering is involved, the short-term memory is naturally consistent and up to date. This pipeline preserves memory while preventing catastrophic overfitting.

%% file: sections/3_experiment.tex
\section{Experiments}
\label{section:experiment}
\subsection{Experimental Settings}
\noindent\textbf{Datsets.} We evaluate our method on three datasets: 
(1) \textit{BernBypass70}~\cite{Lavanchy2024} contains 70 videos averaging 72 minutes. 
%It is evenly split for training/validation and testing.
(2) \textit{Cholec80}~\cite{TwinandaSMMMP17endonet} includes 80 cholecystectomy videos divided into 7 phases, with a mean duration of 39min.
%We use 10/4/7 videos for training/validation/testing. 
(3)  \textit{AutoLaparo}~\cite{wang2022autolaparo} comprises 21 hysterectomy videos, averaging 66 minutes and divided into 7 phases. 
%We use 40/10/20 videos for training/validation/testing. 
All surgical videos are captured at $25$ frames per second (FPS) and subsequently downsampled to $1$ FPS to enable surgeons to accurately annotate specific surgical phases. To maintain experimental consistency, we strictly follow the data splits established in previous studies~\cite{TwinandaSMMMP17endonet,JinD0YQFH18sv-rcnet,JinLCZDH21tmrnet,GaoJLDH21transsvnet,valderrama2020tapir,liu2023skit,Lavanchy2024}. 

% \begin{itemize}[topsep=0pt, wide=0pt]
%     \item \textbf{Cholec80}~\cite{TwinandaSMMMP17endonet} includes 80 cholecystectomy videos divided into 7 phases, with a mean duration of 39min. It is evenly split for training/validation and testing.
%     \item \textbf{AutoLaparo}~\cite{wang2022autolaparo} comprises 21 hysterectomy videos, averaging 66 minutes and divided into 7 phases. We use 10/4/7 videos for training/validation/testing. 
%     \item \textbf{BernBypass70}~\cite{Lavanchy2024} contains 70 videos averaging 72 minutes. We use 40/10/20 videos for training/validation/testing.
% \end{itemize}

%For robust hyperparameter tuning, we carefully select 10 videos from the Cholec80 training set to serve as validation sets. After completing the hyperparameter optimization process, we train our final models using the original training sets.

\noindent\textbf{Evaluation Metrics.} Following the current practice of Cholec80 and AutoLaparo We employ four distinct metrics for surgical phase recognition evaluation: \textit{video-level accuracy}, phase-level \textit{precision}, \textit{recall}, and \textit{Jaccard}.
%The primary metric is \textit{video-level accuracy}, which calculates the mean percentage of correctly recognized frames on a per-video basis. Additionally, to address the inherent class imbalance in surgical phase recognition, we incorporate three phase-level metrics: \textit{precision}, \textit{recall}, and \textit{Jaccard}. Specifically, precision measures the ratio of true positive predictions to total predictions, recall computes the ratio of true positives to total ground truth instances, and the Jaccard represents the ratio of true positives to the union of predictions and ground truths.
As all previous relaxed metric evaluation are based on problematic evaluation code~\cite{funke2023metrics}, we only evaluate with non-relaxed metric. Following SKiT~\cite{liu2023skit}, we first concatenate the predictions and ground truth labels from all videos into a single continuous sequence. We then compute the average performance per phase.
%implementing a specific strategy to handle missing phases in certain videos for phase-level metric calculations. This comprehensive evaluation approach ensures a thorough and fair assessment of our model's performance across different scenarios and datasets. 
For BernBypass70 dataset, we follow its official evaluation protocol~\cite{Lavanchy2024}, including video-level accuracy and phase-level precision, recall, and F1 score. Unlike Cholec80 and AutoLaparo, the phase-level performance metrics are averaged across phases per video and then across videos. %We applied the officially released code for consistency.

\noindent\textbf{Implementation Details.} For the Surgformer~\cite{yang2024surgformer} baseline, we strictly follow their published training and testing protocols using their official implementation. For MoSFormer and alabation study, we use the same hyperparameters as the Surgformer baseline and train for 30 epochs. For history encoding we apply the history step size $30$. For short-term impression, we take 8 impression tokens from intervals of $\{64, 128, 192, 256, 320, 384, 448, 512\}$.

\begin{table}[t]
\centering
\caption{Benchmark results for online surgical phase recognition. Note that, higher numbers reported in the original Surgformer paper~\cite{yang2024surgformer} used a different metric calculation, which has been adapted with our metrics for a fair comparison.}
\resizebox{1.0\linewidth}{!}{
\begin{tabular}{clllll}
\toprule
\multirow{2}{*}{Datasets}  & \multirow{2}{*}{Methods}   & \multirow{2}{*}{\makecell{Video-level\\ Accuracy}} & \multicolumn{3}{c}{Phase-level} \\
\cmidrule(lr){4-6}
&  &   & Precision & Recall & Jaccard/F1 \\
\midrule
& TeCNO ~\cite{czempiel2020tecno} 
& $83.8 \pm 13.6$ & $61.3$ & $62.8$ & $59.2$ \\
& MTMS-TCN ~\cite{Lavanchy2024} 
& $85.3 \pm 13.2$ & $64.6$ & $67.4$ & $62.4$ \\
BernBypass70~\cite{Lavanchy2024}
& Surgformer~\cite{yang2024surgformer}
& $85.9 \pm 12.3$ & $66.1$ & $65.1$ & $62.5$ \\
% &  MoSFormer - L
% & $87.8 \pm 13.6$ \textsubscript{(+1.9)}
% & 68.7 \textsubscript{(+2.6)}
% & 68.1 \textsubscript{(+3.0)}
% & 65.2 \textsubscript{(+2.7)}\\
% &  MoSFormer - S
% & $87.4 \pm 12.5$ \textsubscript{(+1.5)}
% & 70.2 \textsubscript{(+4.1)}
% & 67.8 \textsubscript{(+2.7)}
% & 65.6 \textsubscript{(+3.1)}\\
% &  MoSFormer(ours)
% & \bf{88.0 $\pm$ 13.0 \textsubscript{(+2.1)}}
% & \bf{70.7 \textsubscript{(+4.6)}}
% & \bf{68.7 \textsubscript{(+3.6)}}
% & \bf{66.3 \textsubscript{(+3.8)}}
&  MoSFormer(ours)
& \bf{88.0 $\pm$ 13.0 }
& \bf{70.7 }
& \bf{68.7 }
& \bf{66.3 }
\\
\midrule
& TeSTra~\cite{zhao2022real}
& $90.1 \pm 7.6$ & $82.8$ & $83.8$ & $71.6$ \\
& Trans-SVNet~\cite{GaoJLDH21transsvnet}
& $89.1 \pm 6.6$ & $84.7$ & $83.6$ & $72.5$ \\
& LoViT~\cite{Liu23Lovit}  
& $91.5 \pm 6.1$ & $83.1$ & $86.5$ & $74.2$ \\
Cholec80~\cite{TwinandaSMMMP17endonet}
& SKiT~\cite{liu2023skit}  
& 92.5 $\pm$ 5.1 & $84.6$ & \bf{88.5} & $76.7$ \\
% \hhline{~------}
&  Surgformer~\cite{yang2024surgformer} 
& $92.3 \pm 6.2$ & $87.1$  & $87.6$ & $77.8$ \\
&  MoSFormer(ours)
% & \bf{93.2 $\pm$ 5.4\textsubscript{(+0.9)}}
% & \bf{88.2 \textsubscript{(+1.1)}}
% & 87.8\textsubscript{(+0.2)} 
% & \bf{78.7\textsubscript{(+0.9)}}  \\
& \bf{93.2 $\pm$ 5.4}
& \bf{88.2 }
& 87.8 
& \bf{78.7}  \\
% &  MoSFormer - S
% & 92.4 $\pm$ 5.5\textsubscript{(+0.1)}
% & 87.0  
% & 87.4  
% & 77.4 \\
% &  MoSFormer - M
% & 92.6 $\pm$ 7.0\textsubscript{(+0.3)}
% & 87.9 \textsubscript{(+0.8)} 
% & 87.3  
% & 78.1 \textsubscript{(+0.3)} \\
% \cmidrule(lr){2-8}
\midrule
& TMRNet ~\cite{JinLCZDH21tmrnet} 
& $78.2$ & $66.0$ & $61.5$ & $49.6$ \\
& TeCNO ~\cite{czempiel2020tecno} 
& $77.3$ & $66.9$ & $64.6$ & $50.7$ \\
& Trans-SVNet~\cite{GaoJLDH21transsvnet}
& $78.3$ & $68.0$ & $62.2$ & $50.7$ \\
AutoLaparo~\cite{wang2022autolaparo}
& LoViT~\cite{Liu23Lovit}  
& $81.4 \pm 7.6$ & \bf{85.1} & $65.9$ & $55.9$ \\
& SKiT~\cite{liu2023skit}  
& $82.9 \pm 6.8$ & $81.8$ & $70.1$ & $59.9$ \\
& Surgformer~\cite{yang2024surgformer}  
& $86.1 \pm 7.3$ & $81.5$ & $70.8$ & $62.4$ \\
% &  MoSFormer - L
% & $87.9 \pm 7.3$\textsubscript{(+1.8)}
% & 78.4 
% & \bf{73.8\textsubscript{(+3.0)}}
% & \bf{66.2\textsubscript{(+3.8)}}\\
% &  MoSFormer - S
% & $86.8 \pm 6.9$\textsubscript{(+0.7)} 
% & 77.9
% & 72.6\textsubscript{(+1.8)} 
% & 64.1\textsubscript{(+1.7)} \\
% &  MoSFormer(ours)
% & \bf{88.0 $\pm$ 6.7\textsubscript{(+1.9)}}
% & 84.1\textsubscript{(+2.6)}
% & \bf{73.2}\textsubscript{(+2.4)}
% & \bf{66.2\textsubscript{(+3.8)}}\\
&  MoSFormer(ours)
& \bf{88.0 $\pm$ 6.7}
& 84.1
& \bf{73.2}
& \bf{66.2}\\

\bottomrule
\end{tabular}}
\label{tab:benchmark}
\end{table}

\begin{table}[t]
\centering
\caption{Ablation Study results for online surgical phase recognition. "S", "L", and "M" denotes short-term feature, long-term history, and full memory respectively. "M*" means the counterfactual inference model with ground truth history.}
\resizebox{1.0\linewidth}{!}{
\begin{tabular}{clllll}
\toprule
\multirow{2}{*}{Datasets}  & \multirow{2}{*}{Methods}   & \multirow{2}{*}{\makecell{Video-level\\ Accuracy}} & \multicolumn{3}{c}{Phase-level} \\
\cmidrule(lr){4-6}
&  &   & Precision & Recall & Jaccard/F1 \\
\midrule
% &  Surgformer~\cite{yang2024surgformer} 
% & $92.3 \pm 6.2$ & $87.1$  & $87.6$ & $77.8$ \\
% &  MoSFormer - L
% & \bf{93.2 $\pm$ 5.4\textsubscript{(+0.9)}}
% & \bf{88.2 \textsubscript{(+1.1)}}
% & 87.8\textsubscript{(+0.2)} 
% & \bf{78.7\textsubscript{(+0.9)}}  \\
% &  MoSFormer - S
% & 92.4 $\pm$ 5.5\textsubscript{(+0.1)}
% & 87.0  
% & 87.4  
% & 77.4 \\
% Cholec80~\cite{TwinandaSMMMP17endonet}
% &  MoSFormer - M
% & 92.6 $\pm$ 7.0\textsubscript{(+0.3)}
% & 87.9 \textsubscript{(+0.8)} 
% & 87.3  
% & 78.1 \textsubscript{(+0.3)} \\
% % \cmidrule(lr){2-8}
% \midrule
% & Surgformer~\cite{yang2024surgformer}  
% & $86.1 \pm 7.3$ & $81.5$ & $70.8$ & $62.4$ \\
% &  MoSFormer - L
% & $87.9 \pm 7.3$\textsubscript{(+1.8)}
% & 78.4 
% & \bf{73.8\textsubscript{(+3.0)}}
% & \bf{66.2\textsubscript{(+3.8)}}\\
% AutoLaparo~\cite{wang2022autolaparo}
% &  MoSFormer - S
% & $86.8 \pm 6.9$\textsubscript{(+0.7)} 
% & 77.9
% & 72.6\textsubscript{(+1.8)} 
% & 64.1\textsubscript{(+1.7)} \\
% &  MoSFormer - M
% & \bf{88.0 $\pm$ 6.7\textsubscript{(+1.9)}}
% & 84.1\textsubscript{(+2.6)}
% & 73.2\textsubscript{(+2.4)}
% & \bf{66.2\textsubscript{(+3.8)}}\\
% &  MoSFormer - M*
% & 
% & 
% & 
% & 
% \\
% \midrule
& Surgformer~\cite{yang2024surgformer}
& $85.9 \pm 12.3$ & $66.1$ & $65.1$ & $62.5$ \\
&  MoSFormer - L
& $87.8 \pm 13.6$ \textsubscript{(+1.9)}
& 68.7 \textsubscript{(+2.6)}
& 68.1 \textsubscript{(+3.0)}
& 65.2 \textsubscript{(+2.7)}\\
BernBypass70~\cite{Lavanchy2024}
&  MoSFormer - S
& $87.4 \pm 12.5$ \textsubscript{(+1.5)}
& 70.2 \textsubscript{(+4.1)}
& 67.8 \textsubscript{(+2.7)}
& 65.6 \textsubscript{(+3.1)}\\
&  MoSFormer - M
&  88.0 $\pm$ 13.0 \textsubscript{(+2.1)}
&  70.7 \textsubscript{(+4.6)}
&  68.7 \textsubscript{(+3.6)}
&  66.3 \textsubscript{(+3.8)}
\\
&  MoSFormer - M*
& 89.2 $\pm$ 11.9 \textsubscript{(+3.3)}
& 71.9 \textsubscript{(+5.8)}
& 69.8 \textsubscript{(+4.7)}
& 68.0 \textsubscript{(+5.5)}
\\
\bottomrule
\end{tabular}}
\label{tab:ablation}
\end{table}

\begin{table}[t]
\centering
\caption{Case Study of Counterfactual Inference.}

\resizebox{1.0\linewidth}{!}{
\begin{tabular}{lccccc}
\toprule
\multirow{2}{*}{Video Clip} & Ground & Surgformer& MoSFormer & Erased & MoSFormer \\
& Truth & Prediction & Original History & History & Counterfactual History \\
\midrule
BBP02 3100-3130 & P5 & P1 & P5 & P2,P3,P4,P5 & P2 \\
BBP21 3070-3100 & P8 & P4 & P8 & P5,P6,P7,P8 & P4 \\
BBP29 1330-1360 & P4 & P2 & P4 & P3,P4 & P2 \\
BBP45 1430-1460 & P2 & P1 & P4 & P4 & P2 \\

\bottomrule
\end{tabular}}
\label{tab:counterfactual}
\end{table}

\begin{figure}[t]
  \centering
   \includegraphics[width=1.0\textwidth]{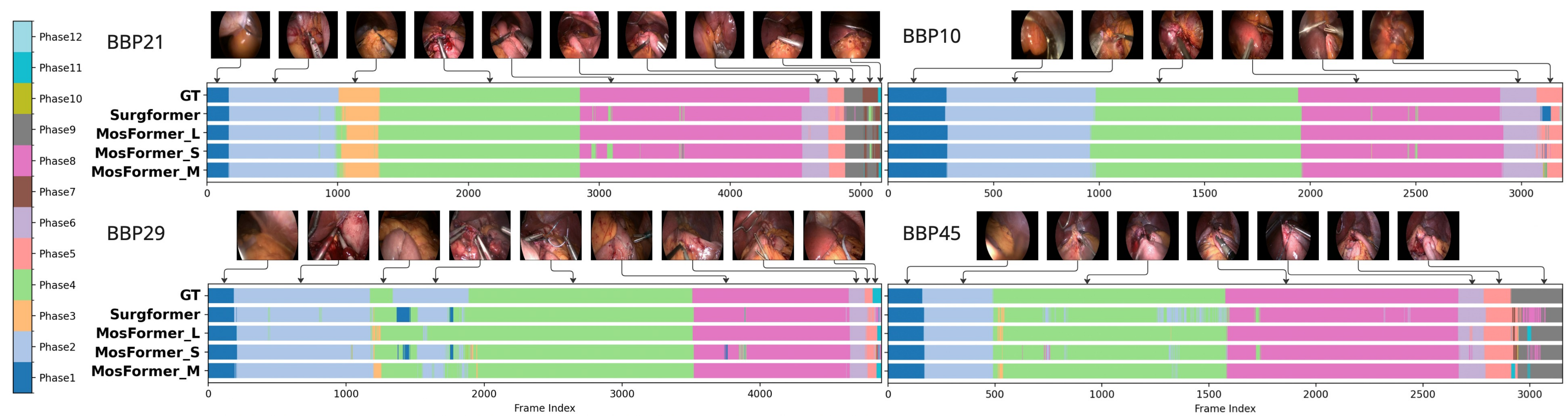}
   \caption{Qualitative Comparision.}
   \label{fig:qualitative}
\end{figure}

\subsection{Benchmark Results:}
\label{sec:benchmark}
We benchmark our approach against state-of-the-art methods for online surgical phase recognition. Results for Surgformer~\cite{yang2024surgformer} are obtained using its official implementation and checkpoints, while the remaining baseline results are taken from published papers or reproduced in~\cite{liu2023skit,Lavanchy2024}. As demonstrated in Table~\ref{tab:benchmark}, MoSFormer achieves state-of-the-art performance on most evaluation metrics across all three benchmarks. Notably, on the more challenging AutoLaparo~\cite{wang2022autolaparo} and BernBypass70~\cite{Lavanchy2024} datasets, MoSFormer significantly outperforms the Surgformer baseline. On AutoLaparo, MoSFormer improves video-level accuracy by 1.9 percentage points (pp), as well as phase-level precision, recall, and Jaccard index by 2.6pp, 2.4pp, and 3.8pp, respectively. On BernBypass70, it achieves a 2.1pp increase in video-level accuracy along with improvements of 4.6pp, 3.6pp, and 3.8pp in phase-level precision, recall, and F1 scores. These promising results validate the effectiveness of MoSFormer architecture.

\subsection{Ablation Study:}
We select BernBypass70~\cite{Lavanchy2024} for the ablation study because it is challenging and has the largest overall scale among the three benchmarks.

\noindent\textbf{Effectiveness of Key Components.} To access effectiveness of key components, We seperately add long-term history and short-term impression to the baseline Surgformer~\cite{yang2024surgformer}, resulting in two variants: MoSFormer-S (short-term impression) and MoSFormer-L (long-term history). Finally, we combine both memory components to form MoSFormer-M, the final version of our model. As shown in Table~\ref{tab:ablation}, incorporating either long-term history or short-term impression individually results in significant performance improvements.
%on the more challenging AutoLaparo~\cite{wang2022autolaparo} and BernBypass70~\cite{Lavanchy2024} benchmarks. 
%On  Cholec80~\cite{TwinandaSMMMP17endonet} the improvement is less significant, we suppose this is because this benchmark is relatively simple and local visual cues are already sufficient to handle most cases. 
These findings indicate that short-term impression and long-term history can be effectively harmonized to form a comprehensive Memory of Surgery representation, thereby advancing the accuracy and robustness of surgical phase recognition. Figure~\ref{fig:qualitative} shows a qualitative comparison between MoSFormer and Surgformer on randomly selected videos. In the illustration, the MoSFormer shows overall better temproal consistency comparing to the Surgformer baseline and less unexpected procedural level mistakes (e.g. Phase 0 after phase 1 in the second and forth video).

\noindent\textbf{Counterfactual Inference.}
To further evaluate the effectiveness of MoS, we leverage the semantic interpretability of the long-term history design and perform counterfactual inference experiments, including a case study and a comprehensive quantitative analysis. First, we intervene on the memory by altering the history representation, which under certain causal models can have a counterfactual inference. For case study, we extract four short video clips, shown in~\ref{fig:qualitative}, where the Surgformer baseline consistently misclassifies the phase. For each clip, we compare prediction results in two scenarios: with real history input (derived from previous predictions) and with intervened history input. Intervened inputs are created by modifying specific history entries, changing their mask values from 1 to 0, effectively erasing the corresponding history phases. The results are summarized in Table~\ref{tab:counterfactual}.
As shown, for the first three rows, both Surgformer and MoSFormer using intervened history generate incorrect predictions. In contrast, MoSFormer with original history correctly identifies the labels, indicating that it leverages the enriched temporal context provided by MoS to resolve ambiguous visual cues. In the last row, MoSFormer initially makes an incorrect prediction due to errors in the original history. When the erroneous history is corrected, its performance improves further.
This is further validated by the quantitative study where we counterfactually feed in ground truth history to the MoSFormer-M model, further improving 1.2 video-level accuracy and 1.7 phase-level F1 score.

%% file: sections/4_conclusion.tex
\section{Conclusion}
\label{section:conclusion}
In this paper, we presented MoS, a complementary approach for integrating augmented temporal context into sliding-window-based ViT architectures for surgical phase recognition. By leveraging procedural-level understanding from short-term impressions and long-term history, MoSFormer advances the state-of-the-art in both accuracy and temporal consistency for surgical phase recognition.
The concept of maintaining and incorporating memory extends beyond surgical phase recognition, offering a generalizable framework for broader surgical video analysis tasks. This approach has the potential to augment the understanding of lengthy, temporally rich procedures. Furthermore, the feature encoding, fusion, and memory caching mechanisms introduced here provide practical techniques for preserving and utilizing video-level context throughout training and inference. The semantic interpretable history design also enables counterfactual inference, facilitating more explainable deployment strategies. Altogether, these contributions pave the way for more comprehensive research in surgical video analysis.